# UHRNet: A Deep Learning-Based Method for Accurate 3D Reconstruction from a Single Fringe-Pattern

Yixiao Wang [a], Canlin Zhou [*a], Xingyang Qi [a], Hui Li [a]

a School of Physics, Shandong University, Jinan 250100, China

* canlinzhou@sdu.edu.cn; phone :+86 13256153609

**Abstract:** The quick and accurate retrieval of an object's height from a single fringe-pattern in Fringe Projection Profilometry has been a topic of ongoing research. While a single-shot fringe-to-depth CNN-based method can restore height map directly from a single pattern, its accuracy is currently inferior to the traditional phase-shifting technique. To improve this method's accuracy, we propose using a U-shaped High-resolution Network (UHRNet). The network uses U-Net's encoding and decoding structure as backbone, with Multi-Level convolution Block and High-resolution Fusion Block applied to extract local features and global features. We also designed a compound loss function by combining Structural Similarity Index Measure Loss (SSIMLoss) function and chunked L2 loss function to improve 3D reconstruction details.

We conducted several experiments to demonstrate the validity and robustness of our proposed method. A few experiments have been conducted to demonstrate the validity and robustness of the proposed method, The average RMSE of 3D reconstruction by our method is only 0.443(mm). which is 41.13% of the U-Net (feature scale=1) method and 33.31% of Wang et al.'s hNet method. Our experimental results show that our proposed method can increase the accuracy of 3D reconstruction from a single fringe-pattern.

## 1. Introduction

Fringe projection profilometry (FPP) is a 3D imaging technique that shows great promise due to its non-contact nature, high spatial resolution, measurement accuracy, and system flexibility. Typically, FPP involves projecting fringe patterns onto an object and capturing the deformed patterns with a camera to reconstruct the object's 3D shape. Various FPP techniques are available, among which phase-shifting profilometry (PSP) stands out for its high accuracy. However, it is not suitable for dynamic scenes due to the requirement of multiple phase-shifting fringes. In contrast, single-shot FPP is fast and efficient as it can acquire 3D information in one shot making it ideal for measuring moving objects or dynamic 3D reconstruction. However, its accuracy is lower than PSP's. Therefore, improving the accuracy of single-shot FPP while maintaining its speed remains an important area of interest in FPP research.

Several solutions have been proposed to overcome these limitations. Wang et al. [1] introduced an enhanced computer-generated Moiré profilometry (CGMP) technique that allows for real-time measurements, albeit with slightly lower accuracy than PSP. Miao et al. [2] combined near-infrared structured light illumination with stereo phase unwrapping to achieve high-precision 3D imaging in real-time. Maeda et al. [3], on the other hand, developed a uniaxial 3D profilometry system using linear polarization patterns and a polarization camera.

Deep learning has recently been used to tackle various challenges in FPP. These include fringe de-noising, phase unwrapping, high-dynamic-range imaging, gamma distortion elimination, and single-shot structure-light 3D imaging. By applying deep learning techniques to FPP, it is possible to improve the accuracy and speed of measurements.

Several studies have proposed the use of deep neural networks for various applications related to fringe analysis. Feng et. al.[4] developed a deep convolutional neural network (CNN) specifically designed for this purpose. Qian et al [5] introduced a deep neural network that utilizes residual connections to perform 3D reconstruction using single-colored fringe images. Wang et al. [6] provided an overview of spatial phase unwrap based on deep learning techniques, while Lin et al.[7] presented an end-to-end network with multistage convolution for structured light fringe denoising[8]. Zhang et al. developed an HDR 3D measurement algorithm that eliminates phase errors caused by HDR. Meanwhile, Suresh et al. proposed the Phase Map Enhancement Network (PMENet) to enhance the quality of FTP phase maps. [9]. Jeught et al. [10] proposed a CNN-based method to directly restore height from a single pattern, while Nguyen et al. [11] studied the effect of using various structured-light patterns on the accuracy of their single-shot fringe-to-depth network. Nguyen et al. [12] proposed the fringe-to-phase network, which directly retrieves three wrapped phase maps from a color fringe pattern. However, this method can only obtain wrapped phase data and requires additional procedures such as phase unwrapping, calibration of the FPP system, and phase-to-height mapping transformation to restore the 3D height map of tested objects using traditional FPP. Although the fringe-to-depth strategy provides straightforward end-to-end processing, research has shown that its performance is inferior to that of fringe-to-phase schemes. Moreover, traditional methods typically have better accuracy than the fringe-to-depth strategy [13]. Van  et.al [14] found that the choice of loss function has a significant impact on accuracy in Structured Light Profilometry based on deep learning. Similarly, Song et al. [15] introduced a network for super-resolution phase retrieval. Deep learning methods have demonstrated better performance in FPP than traditional methods. However, there is still considerable scope for improvement. These studies have motivated us to explore the potential of improving the accuracy of single-shot fringe-to-depth networks.

This paper introduces UHRNet, a new network architecture that aims to enhance the accuracy of the fringe-to-depth network. The proposed UHRNet utilizes the encoding and decoding structure of a U-Net as its backbone. and Multi-Level Convolution Block and High-resolution Fusion Block is applied to extract local features and global features,our idea is inspired by High-Resolution Representation[16] . To further improve the network's accuracy, we developed a more appropriate compound loss function by combining Structural Similarity Index Measure Loss and chunked L2 loss function which is incorporated it into the training process. Our experimental results demonstrate that compared to U-net and h-Net, UHRNet reduces average RMSE by 59.61% and 58.88%, respectively. Overall, our work highlights the potential of UHRNet in significantly improving the accuracy of fringe-to-depth networks for various applications. This paper is organized into four sections. Section 1 provides a review of related work. In Section 2, we describe our proposed network architecture and loss function composition in detail. Section 3 presents experimental and evaluation results. Finally, in Section 4, we conclude by discussing our findings and future work.

2. **Method**

The objective of the proposed method is to enhance the accuracy of the fringe-to-depth network, which transforms a single fringe-pattern into a 3D height map of the object. We incorporated Multi-Level convolution Block, and High-resolution Fusion Block into the network structure. Additionally, we developed a more suitable loss function by combining Structural Similarity Index Measure Loss (SSIMLoss) function and chunked L2 loss. Once the neural network has been trained, we can input the actual fringe pattern for 3D reconstruction. In Section 2.1, we provide an overview of the network structure and its details. In Section 2.2, we explained the compound loss function

**2.1 Network structure**

In the problem of how to use neural network to restore objects from single-frame pattern to 3D height map with high accuracy, the network structure of neural network is the focus of attention. Based on this point, we design an improved network structure based on U-Net and propose UHRNet. The structure of UHRNet is shown in Fig. 1. The main structure of the network adopts the classical encoding and decoding structure of U-Net. Each encoding and decoding layer is composed of Multi-Level convolution Block that can extract information of different depths of the feature map, as shown in Fig. 2. The encoding and decoding layers are linked by High-resolution Fusion Blocks, which are different from the traditional skip connection and they include the fusion layers of multiple scale feature. These High-resolution Fusion Blocks were designed to minimize the loss of high-resolution information during the encoding stage and to effectively integrate information from the feature maps at different depths. The design of the High-resolution Fusion Block was inspired by HRNet's high-resolution representation and its detailed structure is shown in the in Fig. 3.

Because U-Net's encoding-decoding structure has excellent information extraction and generation ability, UHRNet uses it as the backbone structure. As illustrated in Fig. 1, after the input, the single fringe-pattern is transmitted to the encoding layers of the convolutional neural network. The four-layer encoding module progresses layer by layer. The number of channels in the feature map increases gradually to 64C, 128C, 256C, 512C and finally to 1024C (C represents the number of feature map channels, and the number of input pattern channel is 1) after convolution and downsampling. The height and width of the feature map is reduced to 1/2, 1/4, 1/8, 1/16 of the input pattern (the height and width of input pattern is 352 and 640) respectively. The feature map processed by Multi-Level convolution Block at each layer in the encoding and decoding stage are respectively marked as Encoding i and Decoding i (i=1, 2, 3, 4), and High-resolution Fusion Block i($F_i$) is added between Encoding i and Decoding i (i=1, 2, 3) (the feature map in Encoding 4 and Decoding 4 stage are too small in size and resolution to include little high-resolution information, so skip connection is adopted between Encoding 4 and Decoding 4). At the final layer, a 1×1 convolution is used to map each 64C feature map to C feature map.

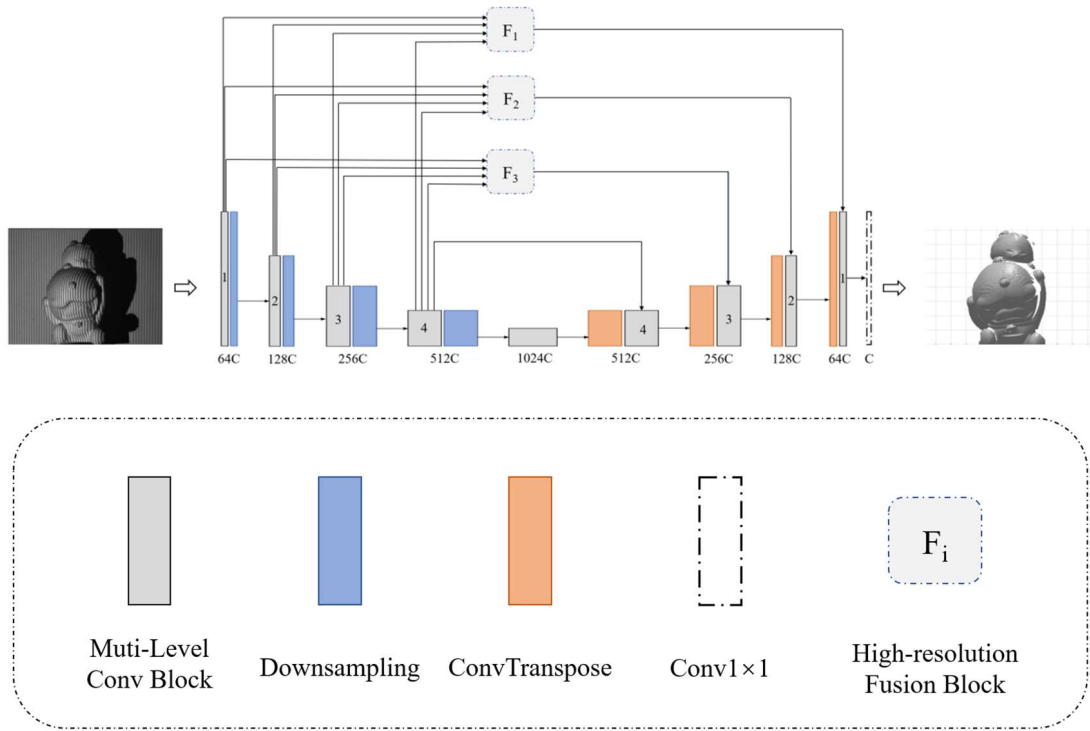

Fig. 1. UHRNet structure

To extract and combine information from input feature maps of varying depths and expand the receptive field of the convolution kernel, we have replaced U-Net's convolution with our designed Multi-Level convolution Block in both encoding and decoding layers. Fig. 2 shows the schematic diagram of the Multi-Level convolution Block. As shown in Fig. 2, the input feature map is divided into five branches. The four branches on the right use dilated convolution with different dilation rates. The dilated convolutions with varying rates can expand the receptive field of the convolutional kernel, allowing for extraction of information from different depths within the feature map. The four branches are merged through concatenation operation. The left branch is a skip connection that is added to the fused feature maps from the other four branches. This ensures smooth propagation of shallow feature information and gradient (each convolution layer is followed by a batch normalization(BN) layer and a Leakrelu activation function which can prevent overfitting, which are not marked in Fig.2).

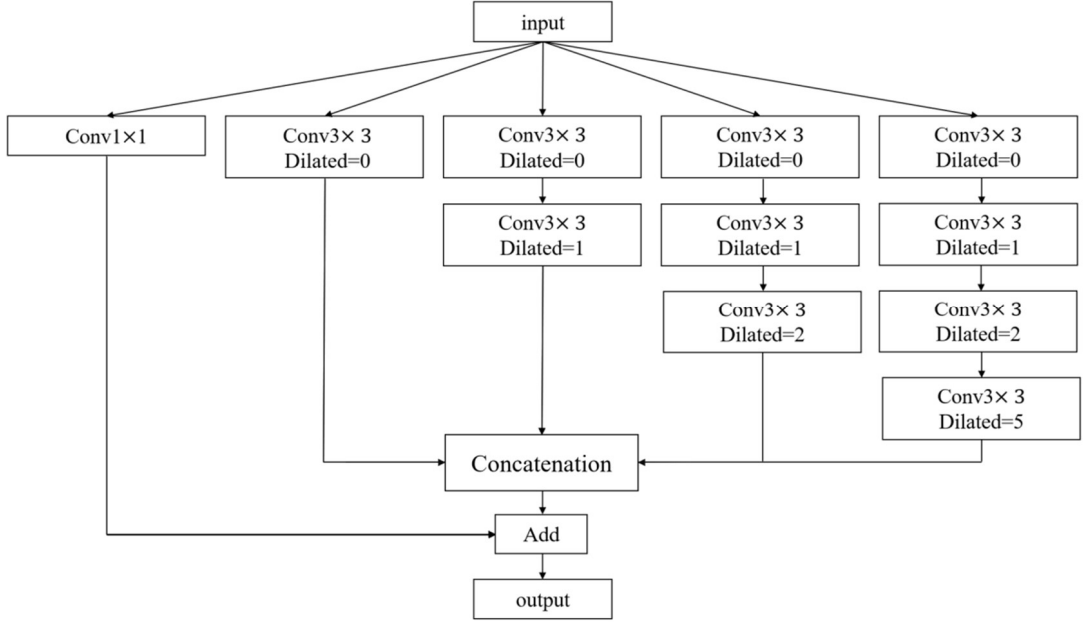

Fig. 2. Multi-Level convolution Block

To enhance the network's ability to reconstruct details, we designed the High-resolution Fusion Block with the aim of retaining as much high-resolution feature from the input feature map. The structure of this block is illustrated in Fig. 3. Because the input feature maps of $F_i$ come from different encoding modules and have different sizes, it is necessary to adjust the size of each feature map after entering the High-resolution Fusion Block for fusion. We will use $F_2$ in Fig.2 as an example to illustrate adjustment process. We take the dimensions of the feature maps in Encoding 2 as the standard size. Since the dimensions of the feature maps in Encoding 1 are larger than the standard size, we use Down-sampling operation to reduce it to the same size as the standard size. Similarly, for Encoding 3 and Encoding 4, we use ConvTranspose2d operation to adjust their dimensions to match with the standard size since their feature map sizes are smaller than the standard size, As shown in Figure 3, the feature maps from encoding modules 2, 3, and 4 are adjusted to the same size. Then, preliminary fusion is performed using concatenation and convolution operations. The resulting output is further fused with Encoding 1 feature map as input2 using concatenation and convolution operations again. As a result, shallow high-resolution information and deep global information are highly fused and transmitted. Therefore, the High-resolution Fusion Block can obtain high-resolution information in the feature map and fuse information from different depths and scales.

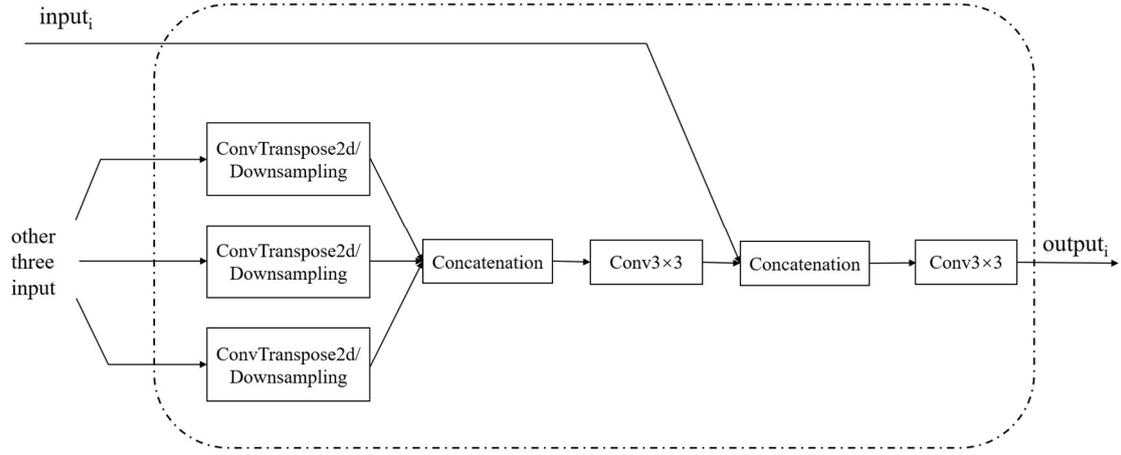

Fig. 3. High-resolution Fusion Block

The Multi-Level convolution Block and High-resolution Fusion Block are crucial elements of UHRNet. Their intricate network structure design empowers UHRNet with precise 3D reconstruction capabilities, as demonstrated in subsequent experiments.

**2.2 compound Loss Function**

The outcome of the network heavily relies on the design of its loss function. We designed a more appropriate loss function to enhance the details of the 3D reconstruction height map. After comparing different experiments, we have chosen the compound Loss function as our preferred loss function. It is composed of two components: chunked L2 loss and SSIM-Loss. The compound loss can be expressed as follows:

$$Fusion\ Loss = chunked\ L2\ Loss + \alpha * SSIMLoss \tag{1}$$

Where $\alpha$ is an adjustable parameter. Through numerous experiments, we continuously adjust $\alpha$ to compare and analyze experimental results. Eventually, we empirically establish for $\alpha = 1000$.

The main purpose of designing chunked L2 loss is to enhance the ability of the network to reconstruct the 3D height map details. Through the observation of the 3D height map generated by the network, we found that the network performance would be excellent in the area where the height map changes slowly. And in areas where the height of the tested object changes abruptly and contains intricate details, the network frequently struggles to accurately reconstruct texture or detail of patterns, resulting in a significant loss of detailed information. Due to the concentration of smooth and detailed areas on the surface of the tested object, we abandon the traditional L2 loss function for the entire pattern and adopt the idea of using chunked L2 loss function. We divide the predicted 3D height map and ground-truth into small patches, calculate the L2 Loss for each patch, and then obtain a weighted sum of all corresponding L2 loss across all patches.

The L2 Loss refers to the mean square error and it can be expressed as:

$$L2\ loss = \frac{1}{n}\sum_{i=1}^{n}(y_i - y_i^p)^2 \tag{2}$$

Where, n is the total number of pixels in the image, $y_i$ is the ground-truth, and $y_i^p$ is the predicted value. Figure 4 illustrates the procedure for computing the chunked L2 loss function. Firstly, the predicted 3D height map and ground-truth are evenly divided into 16 patches. Then the L2 Loss value of each corresponding patch is calculated, and the L2 loss values of various patches are sorted in ascending order. Finally we calculate a weighted sum of all corresponding L2 Losses across all patches according to Eq.(3).

In Fig. 4, the number of patches taken is 16, and it can also be taken as other numbers, which is an empirical data. Generally speaking, it needs to be comprehensively set according to the size of the fringe pattern and the degree of surface height variation of the test object. It should be noted that if the patch area is too large, it will contain both flat and detailed areas, which is not desirable. Conversely, a patch area that is too small will result in an excessive number of patches and unnecessary calculations. In our experiments, we found it appropriate to divide the predicted height map into 16 patches.

After repeated experimental testing, we found that weight design and selection are crucial when calculating weighted L2. The L2 loss value will be smaller if the objects' height in the patch changes slowly and there is less detail. Conversely, it will be larger if there are abrupt changes in object height and more detail present in the patch. Therefore, we assign different weights to the L2 Loss in each patch respectively when calculating the weighted sum of all corresponding local L2 Loss across all patches. Here, we set the weight values like this, that is, the weight of each patch is proportional to the corresponding L2 Loss value, which can be written as eq. (4), We found that selecting weights in this way can produces the good experimental results.

$$chunked\ L2\ Loss = \sum_{i=1}^{16} weight_i \times L2\ loss_i / 16 \qquad (3)$$

$$weight_i = i \times 0.2 - 0.1 \quad (i = 1, 2, 3 \dots 16) \qquad (4)$$

During the network training process, a dynamic reordering mechanism is used for calculating chunked L2 loss. After each epoch, the all patches will be reordered based on their local L2 loss value in the current epoch, the weight of each patch will be dynamically updated accordingly. This dynamic ordering mechanism allows the network to focus more on areas with significant losses in the pattern, which is more conducive to detail reconstruction, and avoids the situation that some positions cannot be trained due to too low weight.

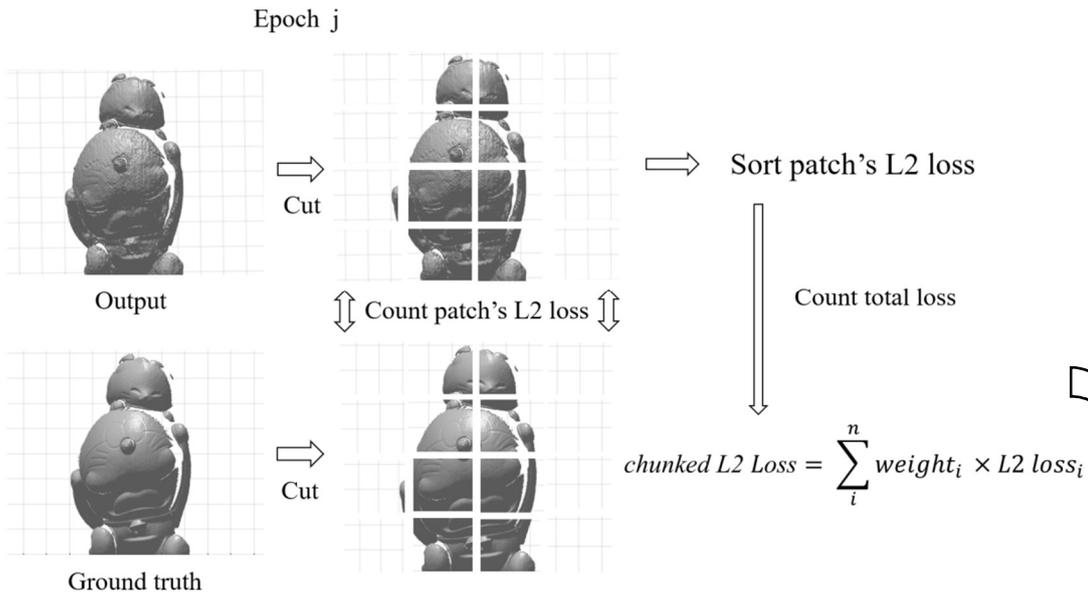

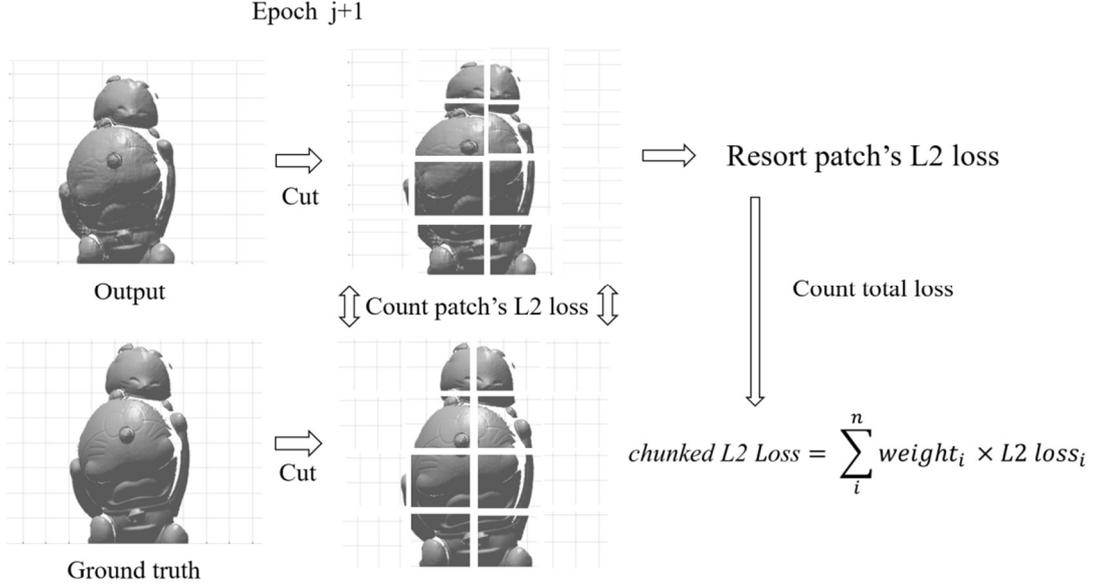

Fig. 4. Calculation flow of *chunked L2 Loss*

Though the L2 loss is easy to train and sensitive to details, it can be easily influenced by anomalies and noise. Moreover, it fails to capture the intuitive feeling of people's visual system when dealing with fringe patterns in FPP. The SSIM function aims to minimize the influence of pattern noise on loss sensitivity and align evaluation criteria with intuitive perceptions. SSIM quantitatively evaluates the similarity of two patterns by brightness l (x,y), contrast c(x,y) and structure s(x,y). The SSIM between two images, x and y, can be calculated using the following formula:

$$S(\mathbf{x}, \mathbf{y}) = f(l(\mathbf{x}, \mathbf{y}), c(\mathbf{x}, \mathbf{y}), s(\mathbf{x}, \mathbf{y})) \tag{5}$$

Where brightness l(x,y), contrast c(x,y) and structure s(x,y) are respectively calculated using the following formulas:

$$l(x, y) = \frac{2\mu_x \mu_y}{\mu_x^2 + \mu_y^2} \tag{6}$$

$$c(x, y) = \frac{2\sigma_x \sigma_y}{\sigma_x^2 + \sigma_y^2} \tag{7}$$

$$s(x, y) = \frac{\sigma_{xy}}{\sigma_x \sigma_y} \tag{8}$$

The SSIM can be expressed as follow:

$$\text{SSIM}(x, y) = \frac{(2\mu_x \mu_y + c_1)(2\sigma_{xy} + c_2)}{(\mu_x^2 + \mu_y^2 + c_1)(\sigma_x^2 + \sigma_y^2 + c_2)} \tag{9}$$

Where, $\mu_x$, $\mu_y$ are the average values of x and y pattern image respectively, $\sigma_x^2$, $\sigma_y^2$ is the variance of the two images, and $\sigma_{xy}$ is the of x and y. $c_1 = (k_1 L)^2$ and $c_2 = (k_2 L)^2$, which are constants used to maintain stability. L is the dynamic range of pixel values, where k1=0.01 and k2=0.03. The value of SSIM is limited between 0 and 1. If the similarity between two pattern images is higher, the SSIM value is closer to 1; otherwise, it is closer to 0. Therefore, *SSIMLoss* can written as:

$$SSIMLoss = 1\text{-}SSIM \tag{10}$$

Using SSIM as the loss function can effectively weaken the sensitivity of the network to outliers and make the generated 3D height map smoother. It is worth noting that since the maximum value of SSIMLoss in a single pattern is 1, it not generally in the same order of magnitude with that of chunked L2 loss. In order for training go smoothly, we weight SSIMLoss to the same order of magnitude as chunked L2 loss. Finally, we adopt the compound loss function, as expressed in equation (1), which can enhance the network's ability to reconstruct details. During the network training process, we use this proposed loss function to assess the discrepancy between the predicted 3D height map and the actual ground-truth.

## 3. Experiments

In this section, we conduct a series of experiments to verify the effectiveness of our method. Section 3.1 is ablation experiment, which demonstrates how various network structures and loss functions affect the outcomes of training (The model is trained and tested using a fringe-pattern dataset consisting of 1532 patterns, and we divided it into training set, validation set and test set according to the ratio of 80%, 10% and 10%). Section 3.2 presents the experimental comparison among hNet, U-Net and our method. The dataset in our experiment are obtained from Wang's article [17], The ground-truth is determined through PSP measurements on plaster sculptures of varying sizes and shapes. To increase the number of samples in the dataset, the sculptures were randomly moved and rotated multiple times. The experiment uses a desktop computer with an Intel Core i7-9700k processor, a 32-GB RAM, and a Nvidia GeForce GTX 2080Ti. The code for training is written in Pytorch and utilizes the Adaptive Moment Estimation (Adam) optimizer. All networks tested in this section were trained for 200 epoches using the training set.

### 3.1 Ablation experiment

In this section, we conduct ablation experiments to confirm the impact of various network structures and loss functions on 3D object reconstruction. This will enable us to determine our final strategy. In this experiment, we use a fringe-pattern dataset as the training set and investigate the impact of 3D object reconstruction on a testing set using four different models, namely A through D. To increase the rigor of the experiment, models A through D exhibit a progressive relationship with only one differing structure between each adjacent model. Model A is u-Net; Model B used Muti-Level convolution Block (MLB) replace U-Net's convolution; Model C added the High-resolution Fusion Block (FB) based on Model B and Fusion Loss is adopted in Model D (Model A-C uses L2 Loss) The structure contained in Model A-D is shown in Table 1:

Table1. Structure composition of Model A-D

| Model | u-Net | MLB | FB | Our loss function |
|---|---|---|---|---|
| A | √ | | | |
| B | √ | √ | | |
| C | √ | √ | √ | |
| D | √ | √ | √ | √ |

After completing the training process, we apply the trained models to test fringe patterns in order to predict their corresponding 3D height map. To evaluate performance, we calculate SSIM and RMSE between the predicted 3D results and ground-truth data. This paper uses two evaluation indexes:

SSIM and RMSE. A value of SSIM closer to 1 indicates a closer match between the model's prediction and the ground-truth. On the other hand, a smaller RMSE index signifies less deviation between the model's prediction and the ground-truth. The RMSE is as follows:

$$\text{RMSE} = \sqrt{\frac{1}{n}\sum_{i=1}^{n}(y_i - y_i^p)^2} \qquad (11)$$

Where, $y_i$ is the ground-truth, $y_i^p$ is the predicted value of the network, and n is the effective number of points.

In order to clearly show the accuracy differences in the 3D reconstruction of the four models, we randomly selected two fringe-patterns in the test dataset, then feed them into the trained network to get the predicted 3D height map outputs, and the results are shown in Fig. 5. Fig. 5(a) and Fig. 5(b) respectively represent the results of two different animal sculptures. In Fig. 5(a), the first row exhibits fringe-pattern and the 3D ground-truth label; each column of the second row presents the 3D height map reconstructed by the Model A, Model B, Model C, Model D respectively. Besides, Four lines of samples are extracted from each restored 3D height map in the same position with the red dotted line marked in the 3D ground-truth label clearly illustrate the height comparison. Fig.5(b) is similar with that of Fig.5(a) except the difference of tested object.

It is evident that Model A can only restore the basic outline of an object, with most of its intricate details being lost. On the other hand, Model B significantly enhances the quality of 3D reconstruction for objects. While Model B can reconstruct most details, the surface of objects may appear rough and certain intricate details cannot be restored. The 3D height map reconstructed by Model C appears smoother and displays more details. This suggests that the High-resolution Fusion Block efficiently transmits high-resolution information from the initial pattern, while minimizing randomness resulting from an increase in feature map size during the decoding stage. Afterwards, we adopted Model D, which utilizes our custom loss function resulting in a further reduction of the difference between the network's prediction results and the ground-truth label. To facilitate a clearer comparison of results, five sample lines are extracted from each restored 3D height map in the same position with the red dotted line marked in the 3D ground-truth label. The height profile along the five sample lines is displayed in the third row of Fig.5(a) and Fig.5(b). Model D exhibits the highest performance in terms of object 3D reconstruction quality and network complexity. Therefore, we have selected Model D as our final method.

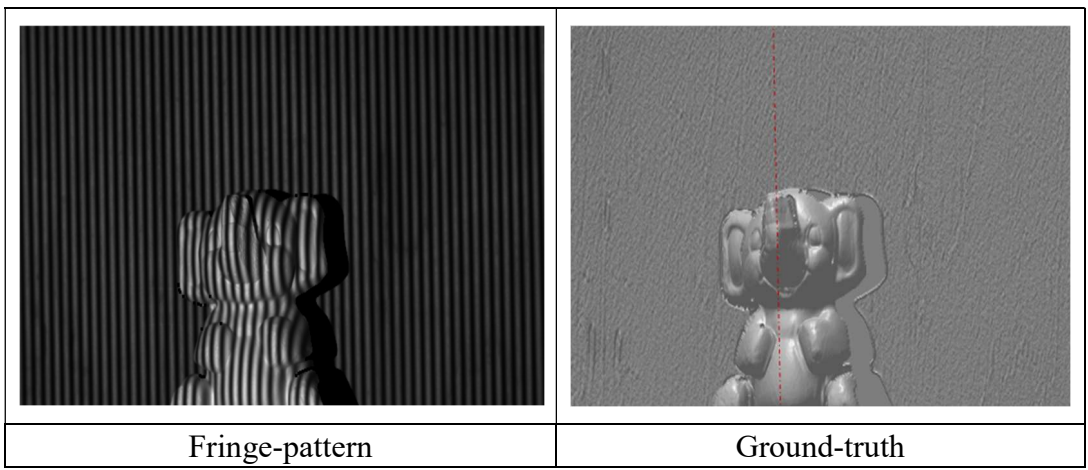

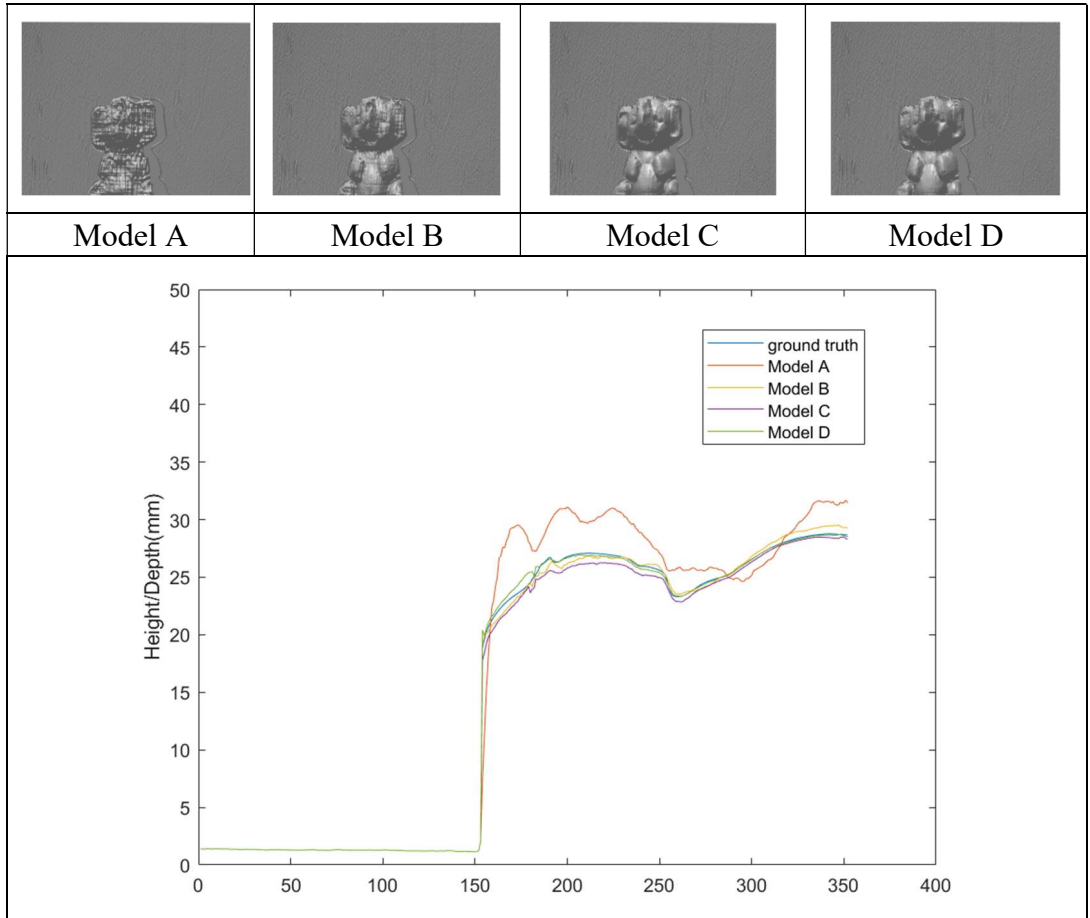

Fig. 5(a). Results of 3-D reconstruction of animal sculptures by different models

The first row: fringe-pattern and the 3D ground-truth label;
Each column of the second row: the 3D height map reconstructed respectively by the Model A, Model B, Model C, Model D
The third row: the height comparison along the same sample line

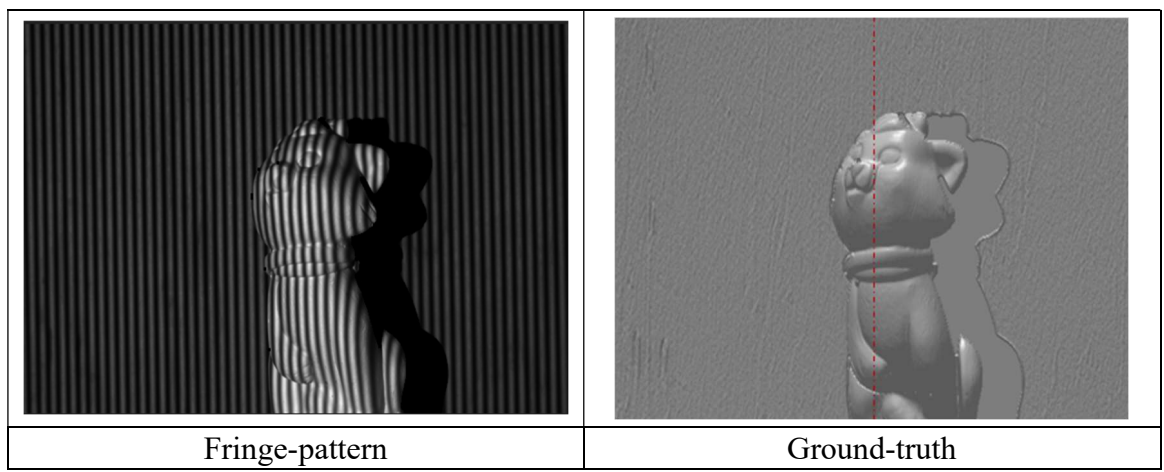

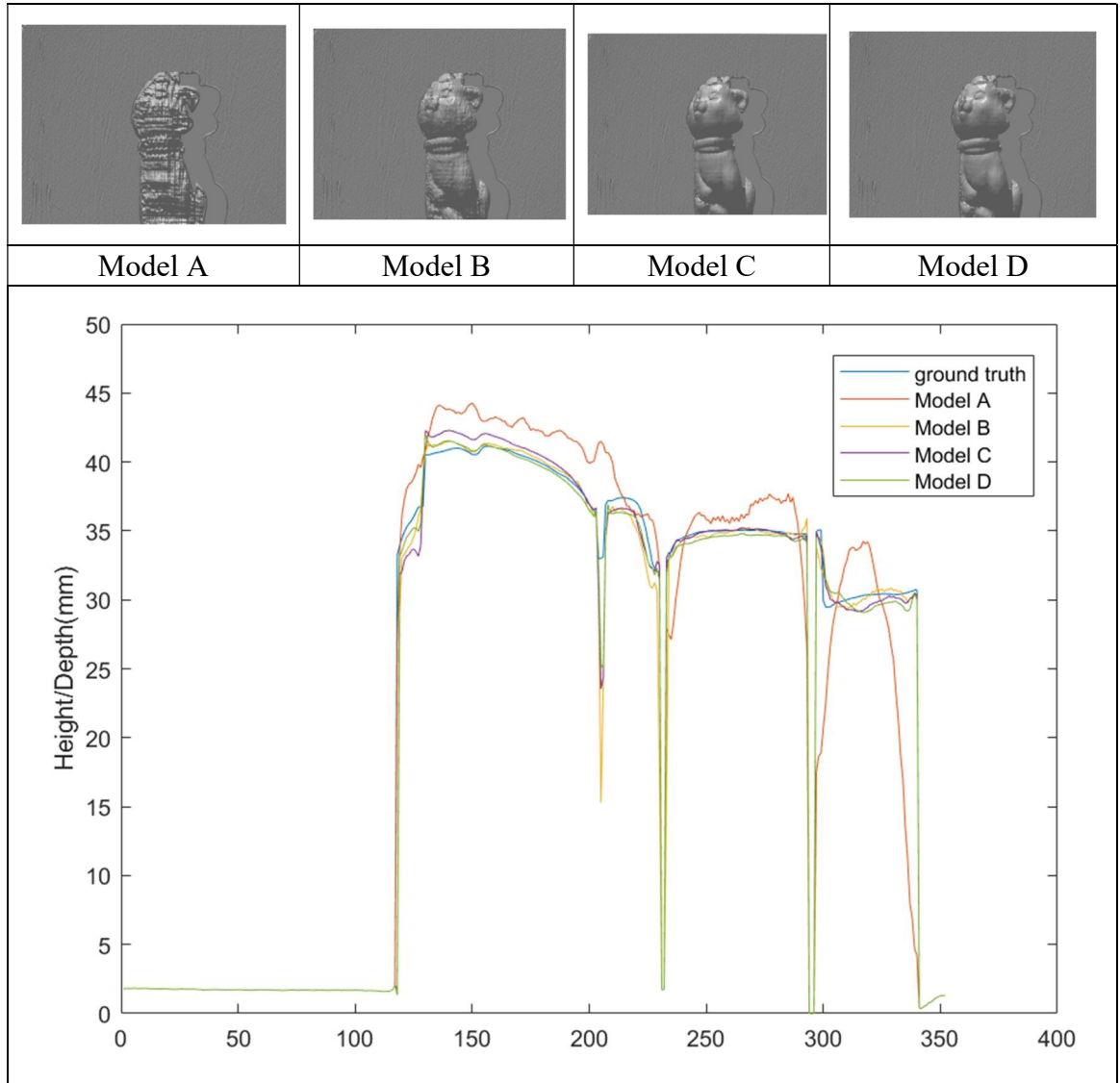

(c)

Fig. 5(b). Results of 3-D reconstruction of another animal sculptures by different models
The first row: fringe-pattern and the 3D ground-truth label;
Each column of the second row: the 3D height map reconstructed respectively by the Model A, Model B, Model C, Model D
The third row: the height comparison along the same sample line

In order to makes quantitative statistics on the 3D height map predicted from model A-D, we calculated the RMSE and SSIM between the results of 3D height map from the four models and ground-truth. The average RMSE and SSIM value in Table 2. are the average values of these two indicators in the test set. First of all, we can see Model A has a small number of parameters but has a large RMSE, the performance of Model B will be greatly improved. Compared with Model A, the average RMSE of the test dataset is reduced by 62.71% and the average SSIM is increased from 0.9655 to 0.9914. Furthermore, Model C will result in an improvement in accuracy. The performance of Model D with compound Loss function will also be further enhanced.

Table 2. Accuracy evaluation of 3D reconstruction of objects

| Model | Param(M) | RMSE(mm) | SSIM |
|---|---|---|---|
| A | **7.76.** | 1.483 | 0.9655 |
| B | 26.11 | 0.553 | 0.9914 |
| C | 30.33 | 0.504 | 0.9957 |
| D | 30.33 | **0.443** | **0.9978** |

**3.2 Experimental comparison between our method and U-Net, hNet**

In this section, several experiments have been carried out to demonstrate the capability and robustness of the proposed method, meanwhile, we compare our method with hNet[11] and U-Net. The performances of the proposed method and hNet and U-Net are first evaluated .Fig. 6 illustrates the history plots of the RMSE and SSIM obtained during the learning process of three models, respectively. The first row is the history curve of the training set in the training stage, The second row is the history curve of the validation set. After about 200 epochs, the RMSE gradually converges to a low value and then tends to be stable. It can be seen that there are obvious RMSE differences among three methods, the SSIM gradually converges to a higher value and then tends to be stable, our method achieved the highest SSIM and lowest RMSE compared to hNet and U-Net. Meanwhile,hNet and U-Net exhibit greater instability during the training process and yields inferior results. The proposed method exhibits faster convergence and achieves better results

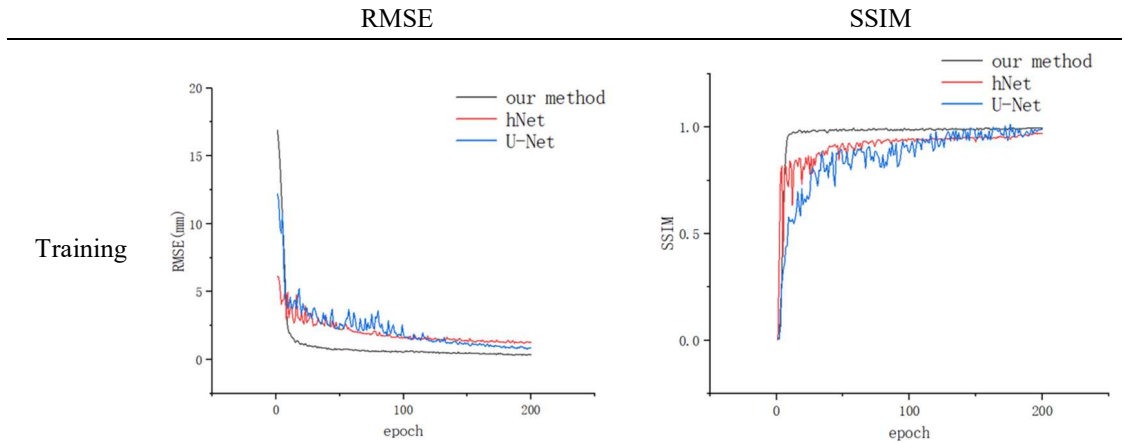

Training

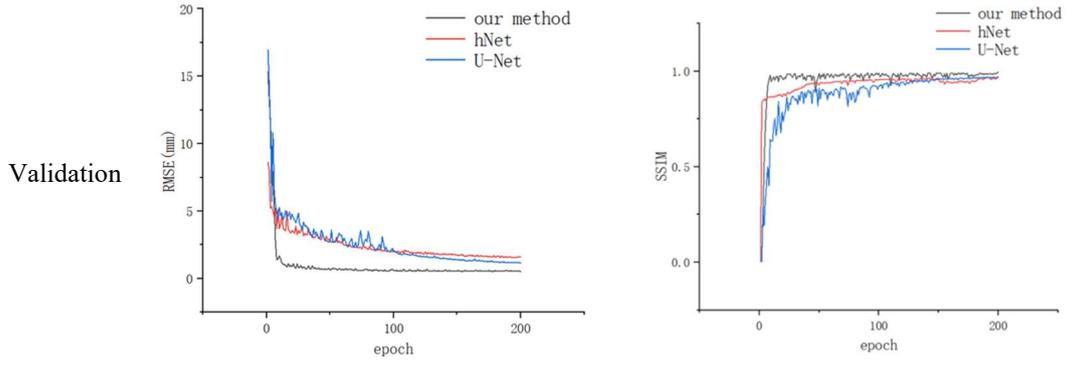

Validation

Fig. 6. RMSE and SSIM plots of our method, hNet and U-Net in learning process
　　　The first column of the first row : the RMSE plots of the training set in the training stage.
　　　The second column of the first row : the SSIM plots of the training set in the training stage.
　　　The first column of the second row : the RMSE plots of the validation set
　　　The second column of the second row : the SSIM plots of the validation set

To quantitatively analyze the reconstruction results of three methods, we chose the ground-truth as a benchmark and calculated the RMSE and SSIM for the three methods. We also measured the inference speed of each network. The results are presented in Table 3. From Table 3, it can be seen that among the three methods, U-Net has the largest number of network parameters, our method achieved an RMSE of 0.443mm, which is 71.41% and 58.87% lower than that of hNet's and U-Net's RMSE, respectively. Additionally, our method achieved a higher SSIM value of 0.9978 compared to hNet's and U-Net 's SSIM value. Therefore, we conclude that our method has significantly higher accuracy than hNet and U-Net. but its SSIM and RMSE are the worst in the final training stage and test set. The UHRNet method has the fewest network parameters, but its SSIM and RMSE are not as good as our method. Our method adopts MLB and FB modules in the network structure to extract different depth and scale information from feature maps for fusion. During training, we also use compound Fusion Loss function. Therefore, our method has the lowest RMSE and highest SSIM. It should be noted that our model's inference speed is slower than the other two model due to the complex network structure. Nevertheless, we believe that it can still perform well in most scenarios.

Table 3. Performance comparison of three methods

|  | RMSE(mm) | SSIM | Parameters(M) | Speed(s) |
| --- | --- | --- | --- | --- |
| Our method | 0.443 | 0.9978 | 30.33 | 0.0224 |
| hNet | 1.330 | 0.9767 | 8.63 | 0.0093 |
| U-Net | 1.077 | 0.9795 | 31.03 | 0.0081 |

We evaluate the performance of the proposed method on some single objects, fringe-pattern of two sculptures are randomly selected from the tested samples, we use hNet, U-Net and our method to predict their 3D height map and results are shown in Fig.7. Fig.7(a) shows the original fringe patterns of the two different sculptures, Fig.7(b) is ground-truths based on PSP, Fig.7(c) shows the predicted 3D results by hNet. Fig.7(d) shows the predicted 3D results by U-Net, Fig.7(e) shows the predicted 3D results by our method. Fig. 7(f) is the comparison of corresponding cross-sections along the red dotted line in Fig.7(b). From Fig.7(c), Fig.7(d) and Fig. 7(e), it can be seen that both three methods can restore the 3D outline, but hNet loses most of the details of the object. The reconstruction effect of U-Net is slightly better than that of hNet, but it still cannot restore the detail of the object satisfactorily. By contrast, our method can retrieve the more details, which can also be seen obviously from the comparison of corresponding cross-sections in 7(f). This illustrates that by the use of Multi-Level convolution Block, High-resolution Fusion Block and compound Fusion Loss, the proposed method can extract more information in a larger spatial range than other two models. To quantitatively analyze the reconstruction results of three methods, we chose the ground-truth as a benchmark and calculated the RMSE and SSIM for the three methods. RMSE and SSIM of our method are 0.440mm, 0.9981, respectively while RMSE and SSIM of hNet are 1.330mm,0.9767, SSIM and RMSE of U-Net are 1.077mm,0.9795.

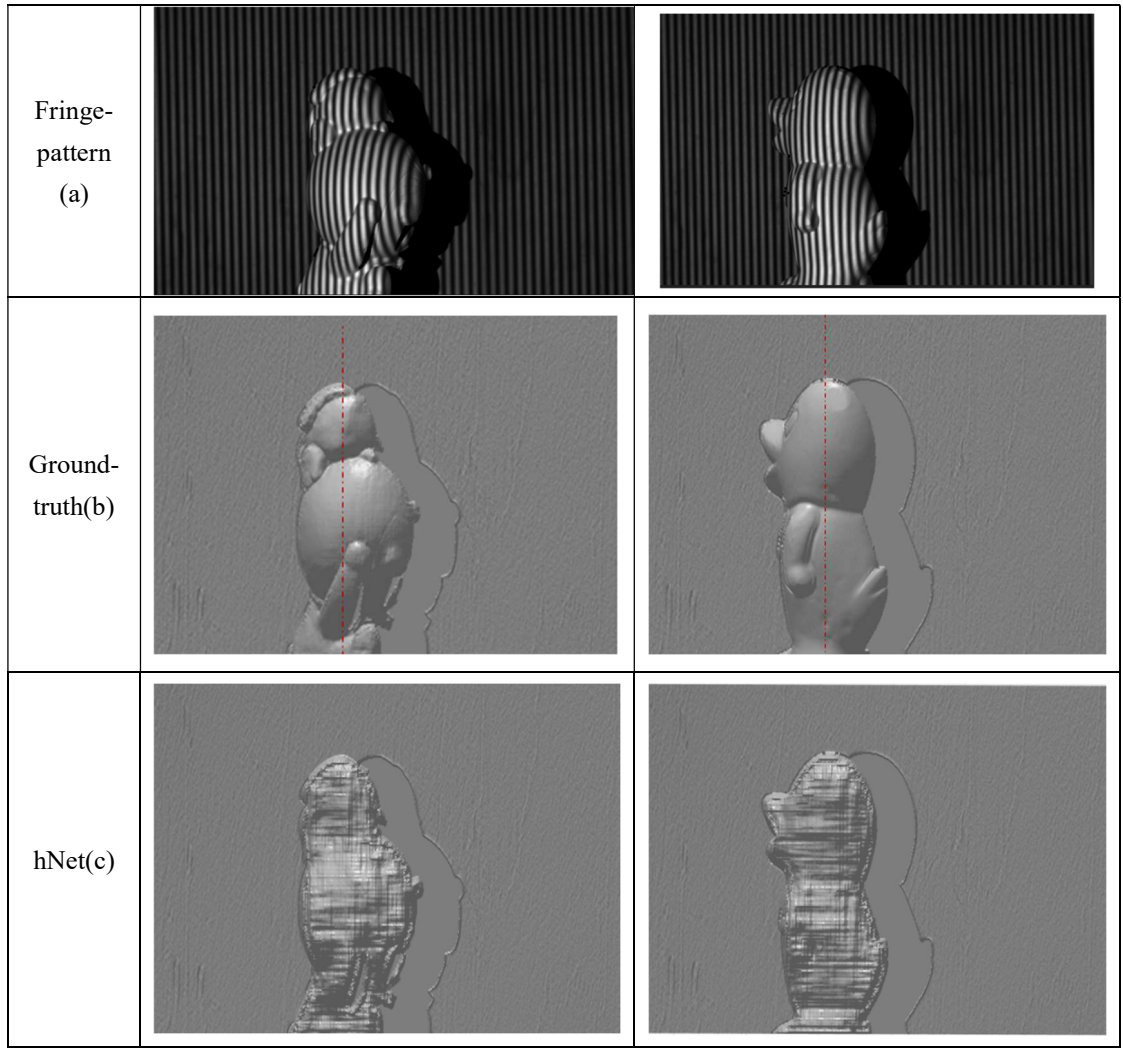

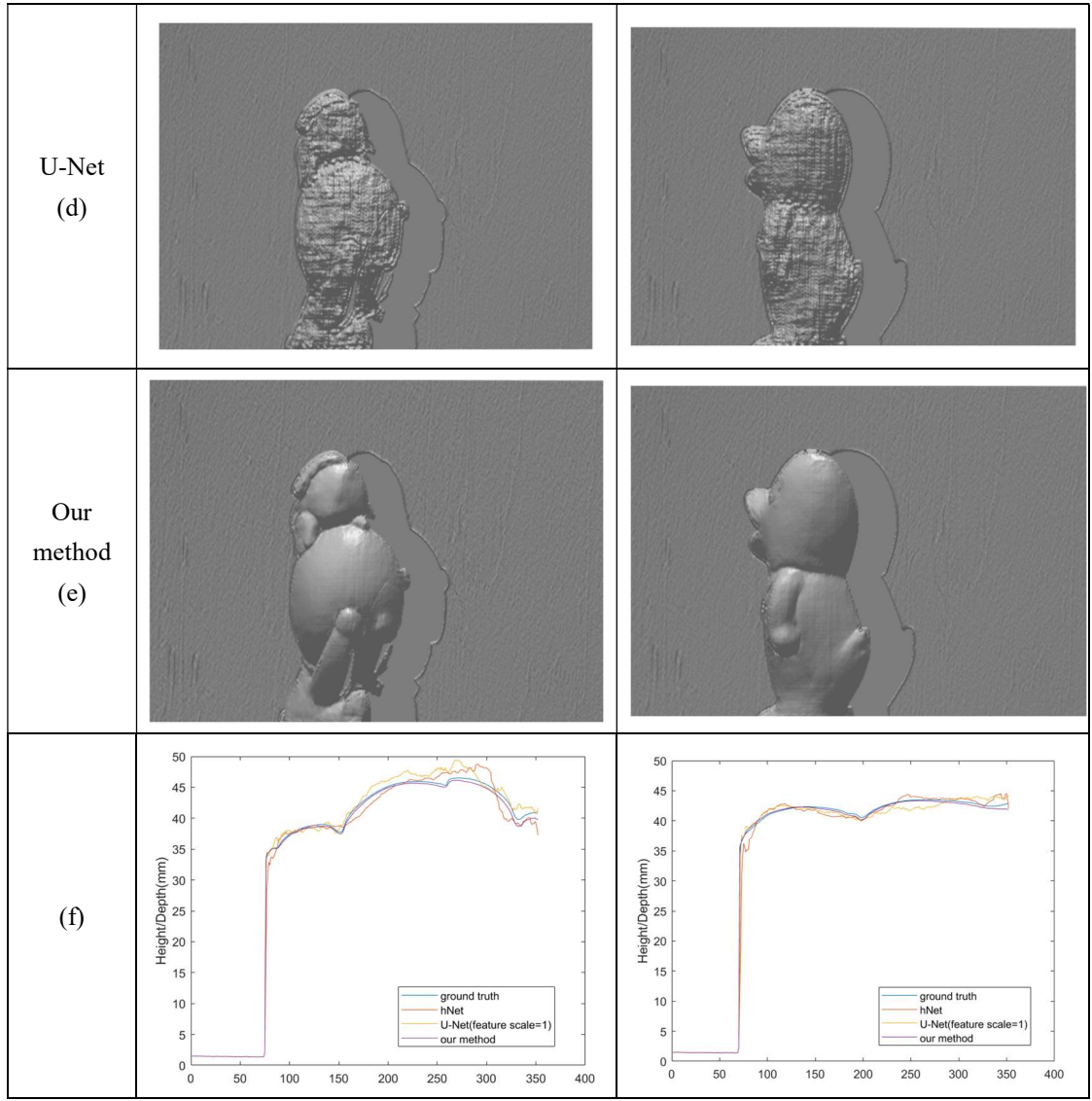

Fig. 7. 3D reconstruction results of two test samples obtained from hNet and our method
(a) the original fringe patterns of the two different sculptures,
(b) the ground-truths,
(c) the predicted 3D results by hNet,
(d) the predicted 3D results by U-Net,
(e) the predicted 3D results by our method,
(f) the comparison of cross-sections along the red dotted line Fig.7(b)

Reconstructing a 3D scene that contains geometric discontinuities, such as multiple separated objects, is a challenge problem for the conventional single-shot fringe projection method. The second experiment is accomplished to validate the proposed method can deal with the multiple separate objects for a single shot 3D shape measurement. In this experiment, two separated objects are placed in the field of view in front of the FPP system. We feed the fringe-pattern captured by camera into the trained networks to get the predicted 3D height map, and the results are shown in Fig. 8. The first row In Fig. 8 exhibits fringe-pattern and the 3D ground-truth label. Each column of the second row presents the 3D height map reconstructed by the hNet, U-Net and our method respectively, the third row presents the height error between ground-truth and 3D height map reconstructed by the three models (the unit of the error map is millimeter). It can be seen that all of the three methods can deal with isolated bodies, and our method's result is close to the ground-truth. As shown in the second row of Fig. 8, hNet only reconstruct general height maps with fewer details, some noisy variations can be observed obviously. The 3D reconstruction result of U-Net is slightly better than that of hNet, but it still can't reconstruct most detail of the object. Compared with the other two, our method can restore the most detail of the object, has highest accuracy. It can also see that our method has the highest accuracy from the height error map as shown in the third row of Fig. 8. We compared the accuracy of three methods in predicting a height map by using the ground-truth as a benchmark. The RMSE and SSIM were calculated to quantitatively analyze their performance. Our method demonstrated superior results with an RMSE of 0.689mm and SSIM of 0.9941, while hNet had an RMSE of 2.079mm and SSIM of 0.9656, and U-Net had an RMSE of 1.613mm and SSIM of 0.9678.

| Fringe-pattern | Ground-truth | |
|---|---|---|
| 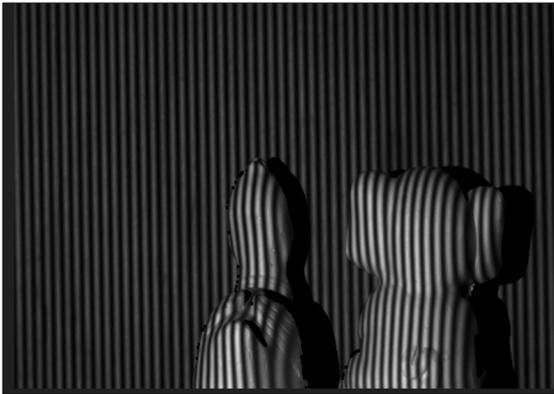 | 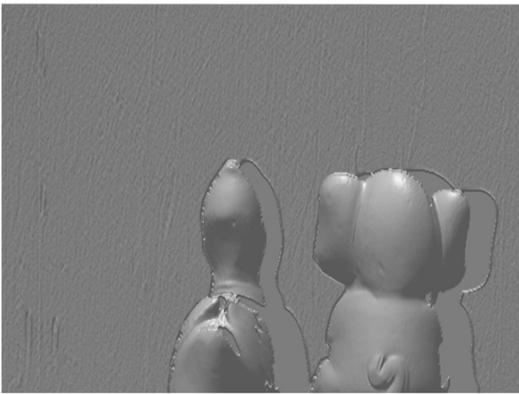 | |
| hNet | U-Net | Our method |
| 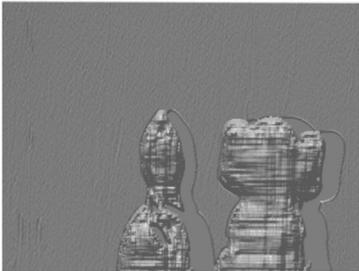 | 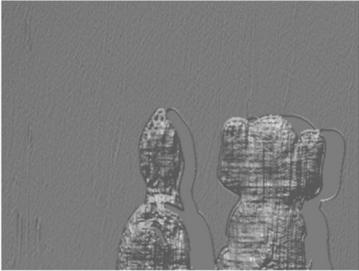 | 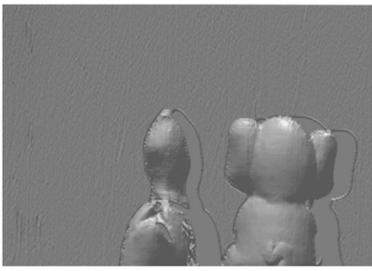 |
| RMSE=2.079(mm) SSIM=0.9656 | RMSE=1.613(mm) SSIM=0.9678 | RMSE=0.689(mm) SSIM=0.9941 |
| height error of  hNet | height error of U-Net | height error of Our method |

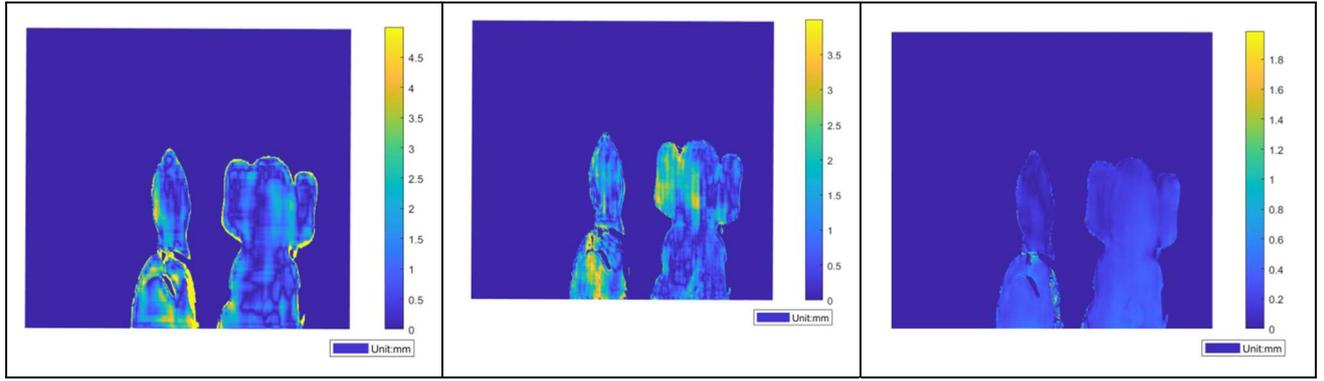

Fig. 8. 3D reconstruction of two isolated objects

The first row: fringe-pattern and the 3D ground-truth label;
The second row: the 3D height map reconstructed respectively by the hNet, U-Net and our method
The third row: the height error between ground-truth and 3D height map reconstructed
respectively by the hNet, U-Net and our method

| Fringe-pattern | Ground-truth | |
|---|---|---|
| hNet | U-Net | Our method |
| RMSE=1.420(mm) | RMSE=1.148(mm) | RMSE=0.576(mm) |
| SSIM=0.9771 | SSIM=0.9788 | SSIM=0.9967 |
| Height error of hNet | Height error of U-Net | Height error of Our method |

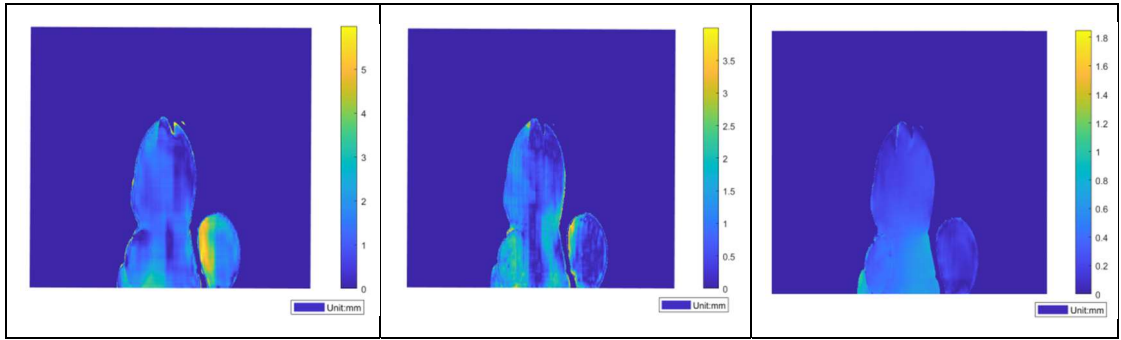

Fig. 9(a). 3D reconstruction of two overlapping object

The first row: fringe-pattern and the 3D ground-truth label;
The second row: the 3D height map reconstructed respectively by the hNet,
U-Net and our method
The third row: the height error between ground-truth and 3D height map reconstructed respectively by the hNet, U-Net and our method

| Fringe-pattern | | Ground-truth |
|---|---|---|
| hNet | U-Net | Our method |
| RMSE=1.849(mm) | RMSE=1.500(mm) | RMSE=0.763(mm) |
| SSIM=0.9613 | SSIM=0.9673 | SSIM=0.9968 |
| height error of  hNet | height error U-Net | height error of Our method |

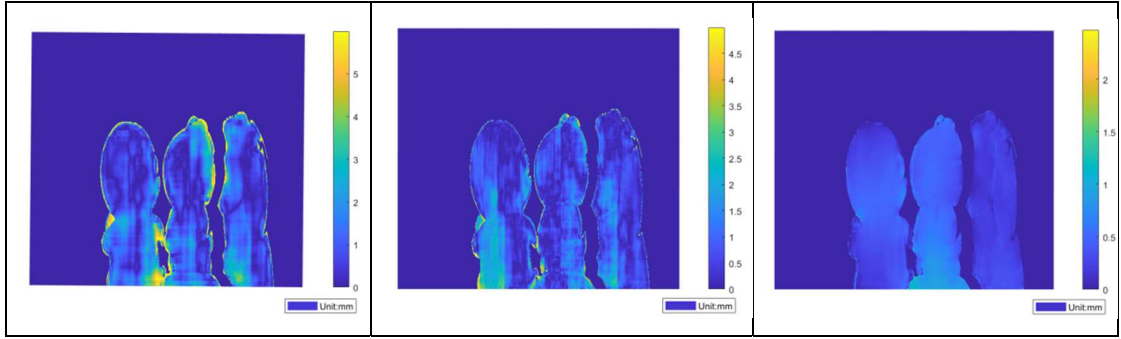

Fig. 9(b). 3D reconstruction of three overlapping object

The first row: fringe-pattern and the 3D ground-truth label;
The second row: the 3D height map reconstructed respectively by the hNet, U-Net and our method
The third row: the height error between ground-truth and 3D height map reconstructed respectively by the hNet, U-Net and our method

At last, we do some experiments on complex multiple composite objects, three separated objects are staggered before and after in the field of view in front of the FPP system, as seen in the first column of the first row in Fig.9. The conventional single shot fringe projection method can't solve this kind of problems well. The experimental procedure is similar to the second experiment. The fringe-pattern is processed by hNet, U-Net and our method. The second column of the first row in Fig.9 is the ground-truth. The second row in Fig.9 shows the height map predicted by hNet, U-Net and our method. By comparing the predicted results shown in the second row of Figure 9, it is evident that our method accurately reconstructs the positional relationships, edge contours, and details of all three objects. However, hNet and Unet fail to precisely reconstruct misaligned parts and intricate details of these objects.

To quantitatively analyze the height map predicted by three methods, we selected the ground-truth as a benchmark and calculated their RMSE and SSIM. Our method outperformed hNet and U-Net with an RMSE of 0.763mm and SSIM of 0.9968, while hNet had an RMSE of 1.8490mm and SSIM of 0.9613, and U-Net had an RMSE of 1.500mm and SSIM of 0.9673.

## 4.  Discussions and conclusion

This paper introduces UHRNet, an enhanced fringe-to-depth network that utilizes Multi-Level convolution Block, High-resolution Fusion Block to incorporate more contextual features for restoring 3D details of the tested object. Additionally, we have developed a compound loss function by combining SSIM and chunked L2 loss which was integrated into the training process. We conducted extensive experiments to evaluate our proposed model and compared it with other methods. Our experimental results demonstrate that compared to hNet and U-Net, UHRNet reduces RMSE by 66.69% and 58.87%, respectively. Our method accurately restores the 3D height map of the tested object, which is comparable to that of the classical phase-shifting method. Compared with the classical phase-shifting method, which needs to capture multiple patterns, the proposed method only requires capturing a single fringe-pattern. It is practical in enormous industrial scenarios where real-time and dynamic 3D measurement is urgent. The proposed technique has some limitations. A network model that has been trained is only valid for a particular geometric configuration. If the camera and projector's relative position changes, it is necessary to retrain the deep-learning model using new dataset under the updated geometric configuration. Of course, transfer learning may partially solve this limitation after the measurement system configuration has changed. In the future, we will research how to restore height map directly from a single pattern based on the untrained deep learning.

**Acknowledgment**: supported by the National Natural Science Foundation of China (Grant no:11672162).

**Disclosures**. The authors declare no conflicts of interest.

**Data availability**. The pytorch code UHRNet will be released at the URL: https://github.com/fead1/UHRNet,The datasets are from reference[17](https://figshare.com/s/c09f17ba357d040331e4).

# REFERENCES

[1] WANG L, CAO Y, LI C, et.al. Improved computer-generated moiré profilometry with flat image calibration[J/OL]. Applied Optics, 2021, 60(5): 1209. DOI:10.1364/AO.412291.

[2] MIAO B, QIAN J, LI Y, et.al. Real-time 3D measurement system based on infrared light projection[C/OL]//EHRET G, CHEN B, HAN S. Optical Metrology and Inspection for Industrial Applications IX. Online Only, China: SPIE, 2022: 35[2023-02-02].

[3] MAEDA Y, SHIBATA S, HAGEN N, et.al. Single shot 3D profilometry by polarization pattern projection[J/OL]. Applied Optics, 2020, 59(6): 1654. DOI:10.1364/AO.382690.

[4] FENG S, CHEN Q, GU G, et.al. Fringe pattern analysis using deep learning[J/OL]. Advanced Photonics, 2019, 1(02): 1. DOI:10.1117/1.AP.1.2.025001.

[5] QIAN J, FENG S, LI Y, et.al. Single-shot absolute 3D shape measurement with deep-learning-based color fringe projection profilometry[J/OL]. Optics Letters, 2020, 45(7): 1842. DOI:10.1364/OL.388994.

[6] WANG K, KEMAO Q, DI J, et.al. Deep learning spatial phase unwrapping: a comparative review[J/OL]. Advanced Photonics Nexus, 2022, 1(01)[2023-02-02].

[7] LIN B, FU S, ZHANG C, et.al. Optical fringe patterns filtering based on multi-stage convolution neural network[J/OL]. Optics and Lasers in Engineering, 2020, 126: 105853. DOI:10.1016/j.optlaseng.2019.105853.

[8] ZHANG L, CHEN Q, ZUO C, et.al. High-speed high dynamic range 3D shape measurement based on deep learning[J/OL]. Optics and Lasers in Engineering, 2020, 134: 106245. DOI:10.1016/j.optlaseng.2020.106245.

[9] SURESH V, ZHENG Y, LI B. PMENet: phase map enhancement for Fourier transform profilometry using deep learning[J/OL]. Measurement Science and Technology, 2021, 32(10): 105001. DOI:10.1088/1361-6501/abf805.

[10] VAN DER JEUGHT S, DIRCKX J J J. Deep neural networks for single shot structured light profilometry[J/OL]. Optics Express, 2019, 27(12): 17091. DOI:10.1364/OE.27.017091.

[11] NGUYEN H, LY K L, TRAN T, et.al. hNet: Single-shot 3D shape reconstruction using structured light and h-shaped global guidance network[J/OL]. Results in Optics, 2021, 4: 100104. DOI:10.1016/j.rio.2021.100104.

[12] NGUYEN A H, SUN B, LI C Q, et.al. Different structured-light patterns in single-shot 2D-to-3D image conversion using deep learning[J/OL]. Applied Optics, 2022, 61(34): 10105. DOI:10.1364/AO.468984.

[13] NGUYEN H, NOVAK E, WANG Z. Accurate 3D reconstruction via fringe-to-phase network[J/OL]. Measurement, 2022, 190: 110663. DOI:10.1016/j.measurement.2021.110663.

[14] VAN DER JEUGHT S, MUYSHONDT P G G, LOBATO I. Optimized loss function in deep learning profilometry for improved prediction performance[J/OL]. Journal of Physics: Photonics, 2021, 3(2): 024014. DOI:10.1088/2515-7647/abf030.

[15] SONG J, LIU K, SOWMYA A, et.al. Super-Resolution Phase Retrieval Network for Single-Pattern Structured Light 3D Imaging[J/OL]. IEEE Transactions on Image Processing, 2023, 32: 537-549. DOI:10.1109/TIP.2022.3230245.

[16]Sun, K., Xiao, B., Liu, D., & Wang, J. (2019). Deep high-resolution representation learning for human pose estimation. In Proceedings of the IEEE/CVF conference on computer vision and pattern recognition (pp. 5693-5703).

[17]H. Nguyen, Y. Wang, and Z. Wang, "Learning-based 3D imaging from single structured-light image" *Graphical Models* 126 (2023): 101171.